\documentclass[11pt]{article}

\usepackage[utf8]{inputenc}
\usepackage[T1]{fontenc}
\usepackage[letterpaper,left=0.88in,right=0.88in,top=0.72in,bottom=0.9in,headheight=22pt]{geometry}
\usepackage{microtype}
\usepackage[dvipsnames]{xcolor}
\usepackage{libertinus}
\usepackage[scale=0.92]{sourcesanspro}
\usepackage{amsmath,amssymb,amsfonts}
\usepackage{bm}
\usepackage{xurl}
\usepackage{url}
\usepackage{graphicx}
\usepackage{float}
\usepackage{tabularx}
\usepackage{booktabs}
\usepackage{array}
\usepackage{capt-of}
\usepackage{caption}
\usepackage{titlesec}
\usepackage{enumitem}
\usepackage[most]{tcolorbox}
\usepackage{fancyhdr}
\usepackage{tikz}
\usepackage{hyperref}

\usepackage{amsmath,amsfonts,bm}

\def\eqref#1{equation~\ref{#1}}
\def\1{\bm{1}}

\DeclareMathAlphabet{\mathsfit}{\encodingdefault}{\sfdefault}{m}{sl}
\SetMathAlphabet{\mathsfit}{bold}{\encodingdefault}{\sfdefault}{bx}{n}

\definecolor{ReportNavy}{HTML}{14213D}
\definecolor{ReportBlue}{HTML}{1D4ED8}
\definecolor{ReportCyan}{HTML}{0891B2}
\definecolor{ReportInk}{HTML}{202124}
\definecolor{ReportMuted}{HTML}{5F6368}
\definecolor{ReportTint}{HTML}{F6F8FB}
\definecolor{ReportRule}{HTML}{D9E2EC}

\hypersetup{
  colorlinks=true,
  linkcolor=ReportBlue,
  citecolor=ReportCyan,
  urlcolor=ReportBlue
}
\usetikzlibrary{arrows.meta,positioning}

\color{ReportInk}

\renewcommand{\headrulewidth}{0.35pt}
\renewcommand{\headrule}{\hbox to\headwidth{\color{ReportRule}\leaders\hrule height \headrulewidth\hfill}}
\fancypagestyle{firstpage}{
  \fancyhf{}
  \renewcommand{\headrulewidth}{0pt}
  \fancyfoot[C]{\sffamily\footnotesize\color{ReportMuted}\thepage}
}

\titleformat{\section}
  {\Large\sffamily\bfseries\color{ReportNavy}}
  {\thesection.}
  {0.55em}
  {}
\titleformat{\subsection}
  {\large\sffamily\bfseries\color{ReportNavy}}
  {\thesubsection.}
  {0.55em}
  {}
\titleformat{\subsubsection}
  {\normalsize\sffamily\bfseries\color{ReportNavy}}
  {\thesubsubsection.}
  {0.55em}
  {}
\titlespacing*{\section}{0pt}{2.1ex plus 0.6ex minus 0.2ex}{1.0ex plus 0.2ex}
\titlespacing*{\subsection}{0pt}{1.6ex plus 0.4ex minus 0.2ex}{0.8ex plus 0.2ex}
\titlespacing*{\subsubsection}{0pt}{1.2ex plus 0.3ex minus 0.1ex}{0.6ex plus 0.1ex}

\setlist[itemize]{leftmargin=1.5em,itemsep=1pt,topsep=3pt,parsep=0pt,partopsep=0pt}
\setlist[enumerate]{leftmargin=1.7em,itemsep=1pt,topsep=3pt,parsep=0pt,partopsep=0pt}

\newcommand{\rcite}[2]{\hyperlink{ref:#1}{#2}}
\newcommand{\tref}[1]{\hyperref[#1]{Table~\ref*{#1}}}
\renewenvironment{abstract}
  {\vspace{1.2em}\begin{tcolorbox}[
    colback=ReportTint,
    colframe=ReportRule,
    boxrule=0.45pt,
    arc=2pt,
    left=13pt,
    right=13pt,
    top=12pt,
    bottom=12pt,
    before skip=0pt,
    after skip=0pt]
    {\sffamily\bfseries\color{ReportNavy}Abstract}\par\vspace{0.55em}\small\linespread{1.04}\selectfont\ignorespaces}
  {\end{tcolorbox}\vspace{1.35em}}

\begin{document}
\thispagestyle{firstpage}
\begin{flushleft}
{\sffamily\fontsize{27}{31}\selectfont\bfseries\color{ReportNavy}Beyond the Black Box:\par}
\vspace{0.08em}
{\sffamily\fontsize{27}{31}\selectfont\bfseries\color{ReportNavy}Interpretability of Agentic AI Tool Use\par}
\vspace{1.05em}
{\sffamily\large\bfseries Hariom Tatsat\textsuperscript{1} \quad Ariye Shater\textsuperscript{1}\par}
\vspace{0.35em}
{\sffamily\normalsize\color{ReportMuted}\textsuperscript{1}Quantitative Analytics, Barclays\par}
\vspace{0.35em}
{\sffamily\small\texttt{hariom.x.tatsat@barclays.com}\quad\texttt{ariye.shater@barclays.com}\par}
\end{flushleft}

\begin{abstract}
AI agents are promising for high-stakes enterprise workflows, but dependable deployment remains limited because these tool-use decisions are difficult to diagnose and control. Agents may skip required tool calls, invoke tools unnecessarily, or take actions whose consequence becomes visible only after execution. Existing observability methods are external: prompts reveal correlations, evaluations score outputs, and logs arrive only after the model has already acted. In long-horizon settings, these failures are costly because an early tool mistake can alter the rest of the execution trajectory, increase token consumption, and create downstream safety and security risk.

We introduce a mechanistic-interpretability toolkit built on \textbf{Sparse Autoencoders (SAEs)}, which decompose activations into sparse internal features, and \textbf{linear probes}, lightweight classifiers that read signals from those features. The framework reads model states before each action and infers whether a tool is needed and how risky the next tool action is. It identifies the model layers and features most associated with tool decisions and tests their functional importance through feature ablation. We train the probes on multi-step agent execution traces from the \textbf{NVIDIA Nemotron function-calling dataset} and apply the same workflow to \textbf{GPT-OSS 20B} and \textbf{Gemma 3 27B} models.

The goal is not to replace external evaluation, but to add a missing layer: visibility into what the model signaled internally before action. This helps surface deeper causes of agent failure, especially in long-horizon runs where an early mistake can impact subsequent agent behavior. More broadly, the paper shows how mechanistic interpretability can support internal observability for monitoring tool calls and risk in agent systems.

\end{abstract}

\par
\vspace{1.25em}

\section{Introduction}
\hypertarget{sec: introduction}{}

An \emph{AI agent} solves a task through repeated decision steps rather than a single response. At each step, it can either answer directly from internal model knowledge or delegate to an external tool, observe the result, and continue. We study this \emph{tool-decision boundary}: the moment when an agent must decide whether to answer directly, call a tool, or take a more consequential external action.

\begin{figure}[t]
\centering
\makebox[\textwidth][c]{\includegraphics[width=1.13\textwidth]{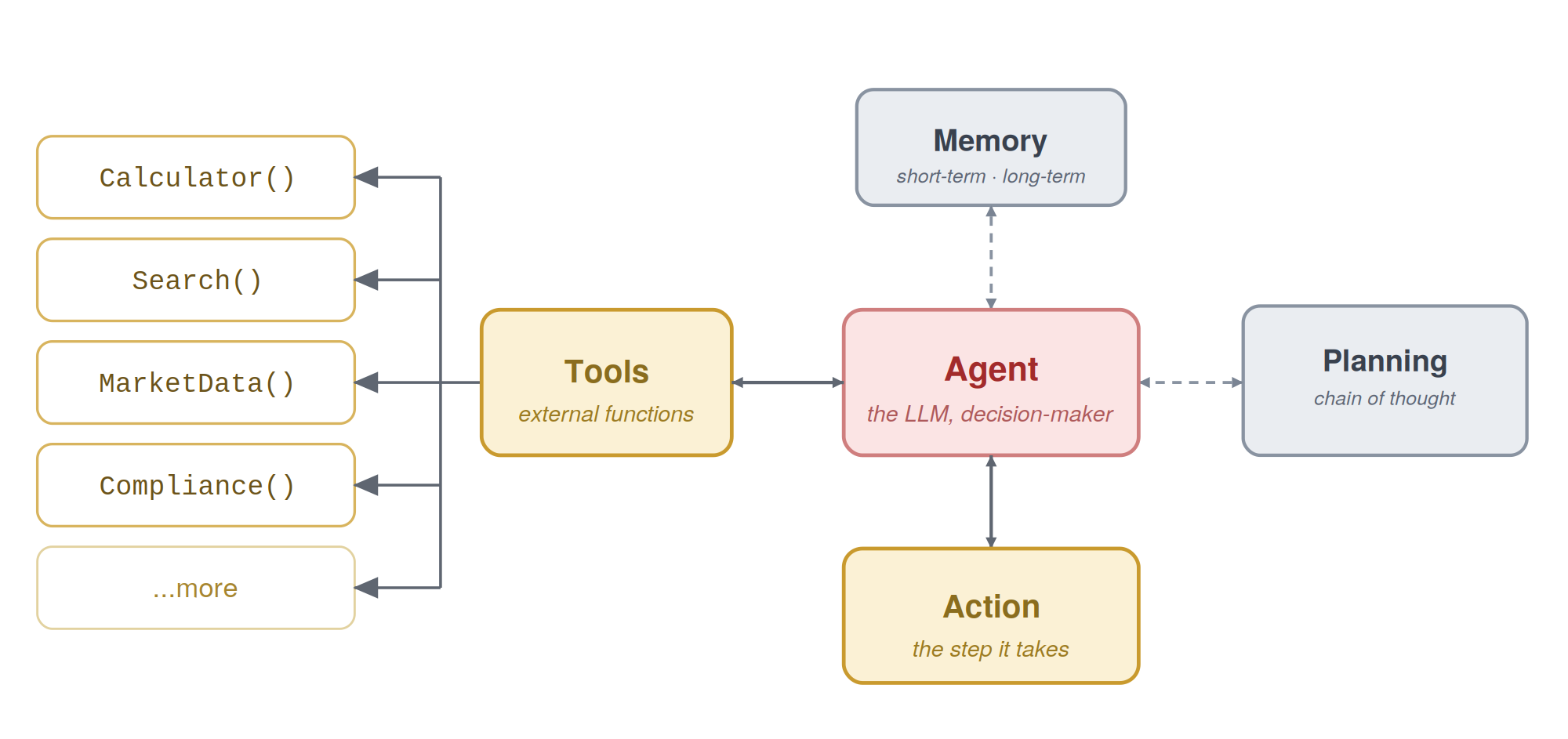}}
\caption{Basic anatomy of an AI agent. A language model acts as the \emph{Agent}, using memory and planning to decide when to call external \emph{tools}; each tool call is an \emph{action}.}
\label{fig: agent_anatomy}
\end{figure}

An AI agent extends a large language model with a few additional capabilities, summarized in Figure~\ref{fig: agent_anatomy}. The model itself acts as the \emph{agent}, the decision-maker that coordinates everything else; while performing core reasoning, the model can draw on \emph{memory} (the conversation so far and information it can look up) and \emph{planning} to choose its next step. Most important for this paper, the model can reason and decide to use \emph{tools}: external functions it calls for what its own knowledge cannot provide, such as downloading latest market data, running a risk calculation, or checking a compliance rule. Each such call is an \emph{action}, and in a high-stakes runtime enviroment, some actions—such as placing a trade or sending a client instruction—carry real and sometimes irreversible consequences. This paper focuses on that tools-and-action pathway, i.e.\ the \emph{tool-decision boundary} introduced above: the moment the agent decides whether to call a tool and how consequential the resulting action is. Today this decision is visible only from the outside, after the call appears in the logs; we instead read the agent's internal state at that moment, before the action fires (Figure~\ref{fig: framework_overview}; formalized in Section~\ref{sec: methodology}).

This boundary is operationally important because tool-use failures are often difficult to diagnose before errors become visible. An agent may skip a required tool call, invoke a tool unnecessarily, or take an action whose consequence becomes visible only after execution. Standard observability methods remain incomplete here. Prompting reveals correlations rather than mechanisms, behavioral evaluation measures outputs rather than internal computation, and logs show what happened only after the model has already acted. The Berkeley Function Calling Leaderboard (BFCL) reflects this limitation in its multi-turn design by combining \textit{state-based} and \textit{response-based} checks, since read-only tool chains may be invisible to state-only evaluation (\rcite{bfcl2025}{Patil et al., 2025}). Related work on tool-selection hallucinations points in the same direction: hidden states can contain useful same-pass signals for tool-call errors that are not visible from outputs alone (\rcite{halluc2026}{Healy et al., 2026}).

We address this gap with an internal monitoring framework that reads model activations immediately before each action and estimates whether the model is internally preparing to delegate to a tool and, secondarily, whether that action is likely to be low, medium, or high risk.

Mechanistic interpretability provides a way to inspect model internals rather than only observe final outputs. Linear probes are lightweight classifiers trained to test whether a concept, such as ``a tool is needed,'' is readable from the model state (\rcite{alain2017}{Alain \& Bengio, 2017}). Sparse Autoencoders (SAEs) go further by decomposing dense activations into sparse internal features that are easier to inspect (\rcite{bricken2023}{Bricken et al., 2023}). We combine both: pre-action activations are first mapped into a sparse feature basis via an SAE, then read out by two task-specific probes: the \textbf{Tool-Need Probe} (binary: tool call vs.\ no tool call) and the \textbf{Tool-Risk Probe} (ternary: low / medium / high risk). This pipeline is lightweight yet interpretable, recovering tool-decision signals from internal state, pinpointing the layers where these signals are strongest, and surfacing the individual features most predictive of tool use and risk. Ablating those features then tests whether the selected sparse features are functionally important for the monitoring readout. Building on our \textit{Beyond the Black Box} line of work (\rcite{tatsat2025}{Tatsat \& Shater, 2025}), this paper extends mechanistic interpretability from static LLM analysis to real-time, pre-action monitoring in multi-step agentic settings.

The choice of SAE features is not only an accuracy choice. A direct linear probe on raw residual activations can test whether tool need is linearly readable, but it leaves the monitor with an opaque direction in dense activation space. SAE features provide a more useful operational object: a score, a layer location, a ranked sparse feature set, and human-inspectable feature labels. This distinction is important in high-stakes enterprise and financial workflows, where monitoring must support audit, escalation, and root-cause analysis rather than only binary prediction. We therefore frame SAE+probe as an agent observability layer: the probe reads the decision signal, while the SAE basis makes the signal inspectable, partially localizable, and easier to compare across failure cases.

The paper makes four contributions:
\begin{itemize}
  \setlength{\itemsep}{1pt}
  \setlength{\parskip}{0pt}
  \setlength{\parsep}{0pt}
  \item A pre-action internal monitoring framework for repeated tool decisions in agent trajectories.
  \item Two complementary readouts: the Tool-Need Probe (Probe~1) and the Tool-Risk Probe (Probe~2).
  \item Localization of tool-decision signals to sparse features and late layers, enabling feature-level inspection and ablation of the monitoring readout.
  \item Evaluation on \textbf{GPT-OSS 20B} and \textbf{Gemma 3 27B instruction-tuned (IT)} models using held-out Nemotron test data and zero-shot BFCL transfer.
\end{itemize}

Section~\ref{sec: related_work} positions the paper relative to prior work and states the research questions. Section~\ref{sec: methodology} defines the decision-point formulation, datasets, internal-state extraction, probe setup, and feature ablation method. Section~\ref{sec: experiments} presents the main empirical results, including held-out Nemotron performance, the illustrative financial trace, layer concentration, and ablation. Section~\ref{sec: runtime_section} evaluates held-out replay and zero-shot BFCL transfer, and Section~\ref{sec: discussion} discusses implications, deployment considerations, and limitations.

\section{Related Work and Research Scope}
\hypertarget{sec: related_work}{}\label{sec: related_work}

Prior work on tool use has mostly evaluated agents from the outside, through end-task success, function-call correctness, or benchmark-specific response scoring. This external evaluation tradition includes learned tool-use setups such as Toolformer and broader function-calling / API benchmarks such as ToolLLM / ToolBench, ToolACE, HammerBench, and BFCL (\rcite{schick2023}{Schick et al., 2023}; \rcite{qin2023}{Qin et al., 2023}; \rcite{liu2024}{Liu et al., 2024}; \rcite{wang2025hammer}{Wang et al., 2025}; \rcite{bfcl2025}{Patil et al., 2025}). BFCL is the most relevant benchmark for our setting because it evaluates abstention and multi-turn behavior, and combines \textit{state-based} and \textit{response-based} checks when read-only tool chains are not visible from final state alone. These benchmarks are essential for measuring observable behavior, but they do not reveal whether the model had internally recognized the need to delegate before acting.

A second line of work studies hidden states directly. Activation-probing results show that internal representations can predict downstream behavior before it is externally visible (\rcite{meco2025}{Li et al., 2025}; \rcite{mckenzie2025}{McKenzie et al., 2025}). In tool-selection settings, internal representations have also been shown to distinguish correct from hallucinated tool calls, with calibration playing an important role when such probes are intended for deployment rather than only offline analysis (\rcite{halluc2026}{Healy et al., 2026}). In parallel, Sparse Autoencoder work shows that dense activations can be decomposed into more interpretable sparse features, making it possible to localize semantically meaningful internal components rather than operate only on opaque residual vectors (\rcite{bricken2023}{Bricken et al., 2023}; \rcite{cho2025}{Cho et al., 2025}).

This distinction motivates a qualitative comparison with simpler internal monitoring alternatives. Output logs and benchmark scores are useful after execution, but they do not expose the internal state before the action. A raw residual-stream probe can test whether a tool-decision signal is linearly readable from dense model activations, and Appendix~\ref{sec: appendix_raw_probe_baseline} shows that this is a strong predictive baseline. However, a dense probe does not by itself provide a sparse, inspectable representation of what triggered the warning. SAE-based probes occupy a different point in this space: they trade some predictive strength for layer localization, sparse feature inspection, feature-level ablation, and more useful evidence for audit and debugging.

\begin{table}[H]
\centering
\scriptsize
\setlength{\tabcolsep}{2pt}
\caption{Qualitative comparison of monitoring approaches. The raw residual probe is the stronger predictive baseline, while SAE+probe is retained for inspectable agent observability.}
\label{tab: qualitative_baselines}
\begin{tabularx}{\linewidth}{@{}p{0.28\linewidth}>{\centering\arraybackslash}p{0.15\linewidth}>{\centering\arraybackslash}p{0.18\linewidth}>{\centering\arraybackslash}p{0.21\linewidth}>{\centering\arraybackslash}p{0.15\linewidth}@{}}
\toprule
\textbf{Approach} & \textbf{Pre-action} & \textbf{Predictive strength} & \textbf{Inspectable features} & \textbf{Audit utility} \\
\midrule
Output/log monitor & No & Limited & No & Low \\
Raw residual probe & Yes & High & No & Medium \\
\textbf{SAE + probe} & Yes & Medium--High & \textbf{Yes} & \textbf{High} \\
\bottomrule
\end{tabularx}
\end{table}

Our work connects these directions in a multi-step agent setting. Rather than evaluating tool use only from outputs, we monitor model state immediately before each action. Rather than probing dense hidden states alone, we probe SAE features that support layer localization, sparse feature inspection, and ablation. This is especially relevant for long-horizon agents, where early tool or coordination failures can propagate through the rest of the trajectory (\rcite{cemri2025}{Cemri et al., 2025}). It is also aligned with the view that external tools should be invoked when they are epistemically necessary, rather than reflexively (\rcite{epistemic2025}{Wang et al., 2025}). It also complements our earlier finance-focused study of mechanistic interpretability, which examined domain-specific LLM behavior rather than pre-action tool decisions in agent trajectories (\rcite{tatsat2025}{Tatsat \& Shater, 2025}).

Framed against Figure~\ref{fig: agent_anatomy}, our questions concern one part of the agent: the moment it decides whether to call a tool and act. The paper is organized around four research questions. \textbf{RQ1} asks whether model activations encode whether a tool should be used at a given decision step—i.e.\ whether the tools-and-action decision of Figure~\ref{fig: agent_anatomy} is present in the internal state before it is externalized as a call. \textbf{RQ2} asks which sparse features and layers most strongly encode tool-need and tool-risk signals. \textbf{RQ3} asks whether internal signals can surface missed and unnecessary tool calls more clearly than logs alone. \textbf{RQ4} asks whether these signals remain useful across repeated decision points and under zero-shot transfer to BFCL. RQ1 and RQ2 are addressed mainly in Sections~\ref{sec: methodology} and~\ref{sec: experiments}; RQ3 and RQ4 are addressed mainly in Sections~\ref{sec: experiments},~\ref{sec: runtime_section}, and~\ref{sec: discussion}.

\section{Problem Formulation and Method}
\hypertarget{sec: methodology}{}\label{sec: methodology}

\begin{figure}[t]
\centering
\includegraphics[width=\textwidth]{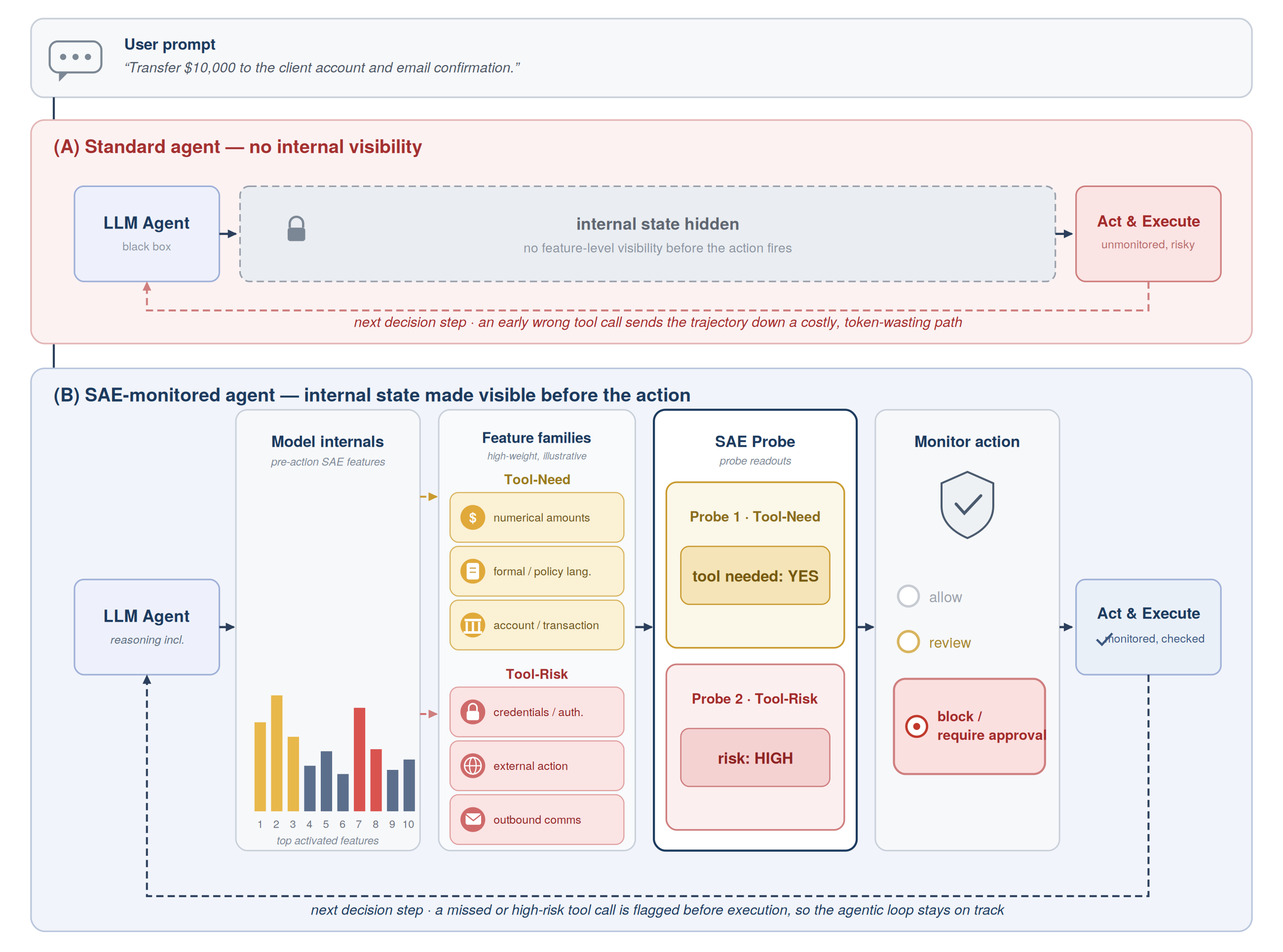}
\captionsetup{font=footnotesize}
\caption{Proposed monitoring framework. \textbf{(A)} A standard agent acts as a black box at the tool-decision boundary. \textbf{(B)} The monitored agent reads pre-action activations through sparse features and probe outputs before mapping the decision to allow, review, or block. Example labels are illustrative; learned features and evaluation appear in Sections~\ref{sec: methodology}--\ref{sec: runtime_section}.}
\label{fig: framework_overview}
\end{figure}

We study agent behavior at repeated \emph{tool decision points}. At each step, we compare three quantities: what the task requires, what the model internally signals, and what the runtime actually does. This three-way view lets us distinguish the main cases that matter for monitoring: correct tool use, missed tool calls, unnecessary tool calls, and high-risk actions. Figure~\ref{fig: framework_overview} contrasts a standard, black-box agent with our proposed SAE-monitored agent and summarizes the full pre-action decision pipeline. Each such decision point is the tools-and-action pathway of Figure~\ref{fig: agent_anatomy}, now examined from inside the model.

Figure~\ref{fig: framework_overview} gives the high-level contrast between a standard tool-using agent and the monitored agent we propose. Both panels follow the same outer loop---Prompt $\to$ LLM Agent $\to$ Act \& Execute, then back to the next decision. In the standard agent, no internal state is read before action, so wrong or missed tool calls become visible only after execution through logs or output checks. In the monitored agent, pre-action activations are decomposed into sparse features, read by Tool-Need and Tool-Risk probes, and mapped to a runtime action such as allow, review, or block.

Panel B sketches what we expect the monitor to read during deployment. The idea is to decompose pre-action activations into sparse SAE features and summarize the strongest features into interpretable \emph{feature families}. The family names shown in the figure---for example numerical, formal-language, account/transaction, credential, external-action, and outbound-communication cues---are representational examples; the actual recovered categories are discussed later in the layer and feature analysis in Section~\ref{sec: results_layers}. These would feed two probe readouts (illustrated here as \emph{tool needed: YES} and \emph{risk: HIGH}), which a monitor maps to a runtime action: allow, review, sandbox, or block / require approval. We envisage, for example, pausing or routing for review when Tool-Need is high but no call is planned, flagging potentially unnecessary tool use when Tool-Need is low but a call is planned, and sandboxing, blocking, or requesting approval when Tool-Risk is high. A central aim of using an SAE basis is that a flagged decision can be traced to specific named features rather than a single opaque score, making it auditable. The feature families and labels in the figure are therefore illustrative; the actual recovered features, automatically generated labels, and layer-level analysis are presented in Section~\ref{sec: results_layers} (Tables~\ref{tab: features_p1}--\ref{tab: features_risk}) and Appendix~\ref{sec: appendix_autointerp}.

Table~\ref{tab: compact_tool_outcomes} summarizes the operational outcomes used throughout the paper. It should be read as the runtime vocabulary for later results: correct behavior, missed delegation, unnecessary delegation, and high-risk action. The Tool-Need Probe provides the internal tool signal, while the Tool-Risk Probe estimates the likely risk tier of the next tool action. These five outcomes are exactly the possibilities at the \emph{Action} step of Figure~\ref{fig: agent_anatomy}: action correctly taken or correctly withheld, a needed action missed, an action taken without need, or a high-consequence action.

\begin{table}[H]
\centering
\footnotesize
\setlength{\tabcolsep}{3pt}
\caption{Runtime outcome taxonomy used throughout the paper. Each outcome compares what the task required, what the agent did, and what the internal probe signaled.}
\label{tab: compact_tool_outcomes}
\begin{tabularx}{\linewidth}{@{}p{0.30\linewidth}X@{}}
\toprule
\textbf{Outcome} & \textbf{Definition} \\
\midrule
Correct no-tool & No tool is required, and no tool is used. \\
Correct tool use & A tool is required, and the runtime calls a tool. \\
Missed tool call & A tool is required, but the runtime does not call one. \\
Unnecessary tool call & No tool is required, but the runtime calls one anyway. \\
High-risk tool call & A tool is used and Probe~2 assigns a high-risk tier. \\
\bottomrule
\end{tabularx}
\end{table}

\subsection{Data preparation}
\hypertarget{sec: dataset}{}\label{sec: dataset}

We convert raw multi-step agent trajectories into per-step decision rows. Each row contains cumulative context truncated at the decision boundary, a binary label indicating whether a tool is required, and a three-level risk label for the next tool action. This preserves a faithful pre-action view: the probe never sees the current step's output or the future trajectory when computing its prediction.

The training data comes from the NVIDIA Nemotron function-calling dataset (\rcite{nemotron2026}{Chandiramani et al., 2026}), where each raw row corresponds to one decision point in a multi-step trajectory. We group rows by trajectory, order them by depth, reconstruct the cumulative context available at each step, and assign a binary \texttt{tool\_needed} label from the gold next action. Tool-call steps are additionally assigned one of three risk tiers: \textbf{low}, \textbf{medium}, or \textbf{high}. Here, risk refers to the likely consequence of the tool action, not the topic domain. Low-risk actions are predominantly read-only retrieval or lookup steps; medium-risk actions involve bounded creation or write operations; high-risk actions include authentication, outbound communication, or dangerous execution actions. Table~\ref{tab: appendix_risk_scheme_summary} in Appendix~\ref{sec: appendix_risk_scheme} summarizes the keyword groups used to instantiate this Nemotron risk-tier scheme.

Probes are trained only on Nemotron-derived step rows. BFCL is reserved for zero-shot transfer evaluation, using the same per-step reconstruction and pre-action probe inference but a different benchmark distribution.

\subsection{Internal state extraction}

We apply the same decision-point pipeline to both backbones: identical per-step context, omission of the current step's generated output from the activation prompt, and layer-wise SAE encoding of pre-action hidden states. For both models, hidden states are mean-pooled over the last 32 pre-action tokens before SAE encoding, rather than read from a single token alone. This choice provides a practical balance between capturing enough immediate context to stabilize the decision signal and keeping activation extraction computationally manageable at runtime.

For \textbf{GPT-OSS 20B}, we read six post-residual layers and encode them with public GPT-OSS SAEs. For \textbf{Gemma 3 27B}, we read four post-block residual layers and encode them with Gemma Scope SAEs. The important point for the paper is not the exact dimensionality of each concatenated vector, but that both models are processed with the same decision-boundary logic and the same probe-based monitoring recipe.

\subsection{Probe training}
\label{sec: probe_training}

The \textbf{Tool-Need Probe} is the primary probe: it predicts whether a tool call is required at the current decision step. The \textbf{Tool-Risk Probe} is secondary: at tool-call steps it predicts whether the next action is low, medium, or high risk. Both probes operate on SAE features rather than raw activations, which makes it possible to inspect layer concentration, identify top sparse features, and test feature necessity through ablation. In terms of Figure~\ref{fig: agent_anatomy}, the Tool-Need Probe reads the agent's decision to take an \emph{action} (call a \emph{tool}), and the Tool-Risk Probe estimates how consequential that action is. 

Formally, let $\tilde{h}^{(\ell)} \in \mathbb{R}^{d}$ denote the pooled pre-action hidden state at layer $\ell$. For each selected layer, a pretrained SAE maps this hidden state to a sparse feature vector
\[
z^{(\ell)}=\phi\!\left(W_{\mathrm{enc}}^{(\ell)}\tilde{h}^{(\ell)}+b_{\mathrm{enc}}^{(\ell)}\right),
\]
where $W_{\mathrm{enc}}^{(\ell)}$ and $b_{\mathrm{enc}}^{(\ell)}$ are the SAE encoder weights and bias for layer $\ell$, and $\phi(\cdot)$ denotes the SAE nonlinearity. We concatenate SAE features across the selected layers,
\[
z=[\,z^{(\ell_1)};\cdots;z^{(\ell_m)}\,],
\]
where $m$ is the number of selected layers. We then fit linear probes on $z$ rather than on raw activations. For Tool-Need, with binary label $y\in\{0,1\}$,
\[
p(y=1\mid z)=\sigma(w^\top z+b),
\]
where $w$ and $b$ are probe parameters and $\sigma(\cdot)$ is the logistic sigmoid. Tool-Risk uses a three-way softmax over $\{\mathrm{low},\mathrm{med},\mathrm{high}\}$. The two probes are trained independently, with distinct targets and feature-ranking criteria, but are evaluated under the same per-step runtime framework.

Each probe is implemented as a sparse logistic classifier over SAE features, with feature selection based on how well each feature separates the target classes and regularization chosen from ridge, lasso, or elastic net. Regularization is applied because the SAE feature space is high-dimensional and often contains correlated latents, so some shrinkage helps control overfitting while keeping the readout interpretable.

To make representative SAE features easier to interpret, we apply an automated feature-labeling step to a small number of selected features. In this workflow, top-activating examples are summarized into short natural-language descriptions using an LLM. More details are deferred to Appendix~\ref{sec: appendix_autointerp}.

\subsection{Feature ranking and ablation}

To test whether top sparse features are merely correlated with probe predictions or functionally important for the monitor, we perform representational ablation directly in SAE feature space. After encoding a step into sparse features, we select a small set of highly ranked SAE features, set them to zero, re-run the probe, and compare the new prediction with the original prediction. If suppressing a small set of latents sharply reduces probe confidence or flips the label, those features are important for the probe's readout. This experiment should be interpreted as evidence about the monitoring mechanism, not as full causal mediation of the model's own tool-call decision. Establishing whether these features causally control the model's generated tool action would require generation-time interventions, which we leave for future work.

\subsection{Evaluation metrics}

We report Tool-Need accuracy, precision, recall, and F1; Tool-Risk accuracy and macro-F1, which weights each risk tier equally; and runtime alignment between expected labels, internal probe decisions, and actual execution. We also report missed-tool warning rates, unnecessary-call warning rates, and risk alerts in replay and transfer settings.

\section{Experiments and Results}
\hypertarget{sec: experiments}{}\label{sec: experiments}

Table~\ref{tab: main_results} gives the headline held-out Nemotron results across both models. The key result is that the Tool-Need Probe (Probe~1) provides the stronger and more stable signal at the tool-call boundary, while the Tool-Risk Probe (Probe~2) becomes useful once a tool call is warranted, where it helps distinguish lower from higher risk actions but is more sensitive to risk-class structure and transfer setting. As described in \hyperref[sec: probe_training]{Section~\ref*{sec: methodology}, Probe training}, the number of selected SAE features varies across probes: GPT-OSS Probe~1 uses lasso with 200 features, all other probes use elastic net, Gemma Probe~1 uses 2000 features, and Probe~2 uses 1000 features for both models.

\begin{table}[H]
\centering
\scriptsize
\setlength{\tabcolsep}{3pt}
\caption{Main held-out Nemotron results. Probe~1 predicts whether a tool is needed; Probe~2 predicts the risk tier of the next tool action.}
\label{tab: main_results}
\begin{tabularx}{\columnwidth}{@{}>{\raggedright\arraybackslash}p{0.46\linewidth}>{\centering\arraybackslash}p{0.23\linewidth}>{\centering\arraybackslash}p{0.23\linewidth}@{}}
\toprule
\textbf{Metric} & \textbf{GPT-OSS 20B} & \textbf{Gemma 3 27B} \\
\midrule
Tool-Need accuracy & 75.3\% & 71.4\% \\
Tool-Need F1 (macro) & 0.75 & 0.71 \\
Tool-Risk accuracy (tool rows) & 90.3\% & 88.5\% \\
Tool-Risk macro-F1 & 0.64 & 0.62 \\
\bottomrule
\end{tabularx}
\end{table}
\vspace{-0.8em}

Appendix~\ref{sec: appendix_raw_probe_baseline} reports a raw residual-stream probe baseline using the same Tool-Need and Tool-Risk formulations. On held-out Nemotron, raw residual probes outperform SAE-feature probes on predictive metrics for both GPT-OSS 20B and Gemma 3 27B IT. We therefore do not position SAE+probe as the accuracy-maximizing choice. The main text focuses on SAE-based monitoring because it provides feature-level evidence, layer localization, and inspectable signals that are useful for audit, escalation, and root-cause analysis in high-stakes agent workflows.

\subsection{Tool-Need Prediction Results}
\hypertarget{sec: results_p1}{}

We first assess whether the model's pre-action internal state contains a reliable signal that an external tool is required. Both models contain a readable tool-decision signal in their SAE features: before the model acts, the internal state often already indicates whether a tool should be used, as summarized in Table~\ref{tab: p1_confusion}.

\begin{table}[H]
\centering
\scriptsize
\setlength{\tabcolsep}{4pt}
\caption{Tool-Need Probe confusion matrices on held-out Nemotron test. Rows are true labels and columns are probe predictions; off-diagonal entries are missed or unnecessary tool-call predictions.}
\label{tab: p1_confusion}
\begin{tabular}{@{}c@{\hspace{2.2em}}c@{}}
\begin{tabular}{@{}lcc@{}}
\toprule
\multicolumn{3}{c}{\textbf{GPT-OSS} (1,993 rows)} \\
 & Pred: 0 & Pred: 1 \\
\midrule
True: no\_tool & 741 & 225 \\
True: tool & 267 & 760 \\
\bottomrule
\end{tabular}
&
\begin{tabular}{@{}lcc@{}}
\toprule
\multicolumn{3}{c}{\textbf{Gemma} (1,821 rows)} \\
 & Pred: 0 & Pred: 1 \\
\midrule
True: no\_tool & 610 & 284 \\
True: tool & 236 & 691 \\
\bottomrule
\end{tabular}
\end{tabular}
\end{table}

GPT-OSS achieves \textbf{75.3\%} accuracy on the held-out Nemotron test, while Gemma achieves \textbf{71.4\%}, as shown in Table~\ref{tab: main_results}. Tool-Need errors are split across both directions rather than collapsing to a single majority class. For GPT-OSS, 760 tool-required steps are correctly identified and 267 are missed; 741 no-tool steps are correctly rejected and 225 are false tool alerts. Gemma shows the same pattern with 691 correct tool predictions, 236 missed tool steps, 610 correct no-tool predictions, and 284 false alerts. This tool-need signal is the paper's primary contribution: it is recoverable with compact feature sets, interpretable in layer space, and later transfers as an omission-auditing signal in runtime settings.

\subsection{Tool-Risk Prediction Results}
\hypertarget{sec: results_p2}{}

We next assess whether the same pre-action state contains information about the likely consequence of the next tool action. Knowing that a tool should be called is necessary but not sufficient: tool actions can differ substantially in risk even when both are valid. Probe~2 therefore asks whether internal representations encode not only the decision to call a tool, but also the likely consequence of the next external action, with the held-out confusion matrices shown in Table~\ref{tab: p2_confusion}.

\begin{table}[H]
\centering
\scriptsize
\setlength{\tabcolsep}{4pt}
\caption{Tool-Risk Probe confusion matrices on held-out Nemotron data. Rows = true tier; columns = predicted tier.}
\label{tab: p2_confusion}
\begin{tabular}{@{}c@{\hspace{2.2em}}c@{}}
\begin{tabular}{@{}lrrr@{}}
\toprule
\multicolumn{4}{c}{\textbf{GPT-OSS} (987 rows)} \\
 & Low & Med & High \\
\midrule
True: low & 818 & 40 & 18 \\
True: med & 16 & 24 & 2 \\
True: high & 17 & 3 & 49 \\
\bottomrule
\end{tabular}
&
\begin{tabular}{@{}lrrr@{}}
\toprule
\multicolumn{4}{c}{\textbf{Gemma} (1,004 rows)} \\
 & Low & Med & High \\
\midrule
True: low & 807 & 35 & 28 \\
True: med & 22 & 21 & 2 \\
True: high & 21 & 7 & 61 \\
\bottomrule
\end{tabular}
\end{tabular}
\end{table}

Both models show strong held-out Nemotron accuracy: \textbf{90.3\%} for GPT-OSS and \textbf{88.5\%} for Gemma on tool-call rows. As shown in Table~\ref{tab: p2_confusion}, the confusion matrices explain why macro-F1 is lower: low-risk cases dominate and are recovered strongly, while medium-risk examples are much fewer and harder to separate. For example, GPT-OSS correctly identifies 818 low-risk rows but only 24 medium-risk rows; Gemma shows a similar pattern with 807 low-risk and 21 medium-risk correct predictions. This asymmetry is consistent across both models, suggesting it reflects the risk-tier structure rather than a model-specific artifact. Probe~2 is therefore useful for risk screening once a tool call is in play, but its medium-risk boundary remains weaker.

\subsection{Illustrative Financial Tool-Use Trace}
\hypertarget{sec: results_finance}{}\label{sec: results_finance}

We retain one illustrative financial tool execution trajectory from the Nemotron distribution (\texttt{trajectory\_id}~3344) to show how the two probes evolve across repeated decision points. This subsection is intended as a worked trace rather than as a standalone evaluation. The quantitative evidence for the paper remains the held-out metrics, confusion matrices, feature tables, and ablation results; the full step-by-step trace and the corresponding plot appear in Appendix~\ref{sec: appendix_qualitative}, specifically Figure~\ref{fig: appendix_finance_probe1} and Table~\ref{tab: appendix_nemotron_financial_full}.

Each step in this trajectory is one instance of the tools-and-action decision of Figure~\ref{fig: agent_anatomy}, observed from inside the model. The main value of this example is that the \textbf{Tool-Need Probe (Probe~1)} rises on steps that genuinely require external financial retrieval and falls on follow-up turns where no new tool call is needed, even though the discussion remains financial. The \textbf{Tool-Risk Probe (Probe~2)} stays predominantly low on those retrieval steps, which shows that the risk signal is not redundant with tool need.

\begin{table}[H]
\centering
\scriptsize
\caption{Illustrative, non-evaluative steps from one Nemotron financial trajectory. $p_{\mathrm{tool}}$ is the Tool-Need probability; Probe~2 probabilities are shown as $(p_{\mathrm{low}}, p_{\mathrm{med}}, p_{\mathrm{high}})$.}
\label{tab: financial_example_main}
\begin{tabularx}{\linewidth}{@{}p{0.09\linewidth}>{\raggedright\arraybackslash}X>{\centering\arraybackslash}p{0.15\linewidth}>{\raggedright\arraybackslash}p{0.32\linewidth}@{}}
\toprule
Step & Summarized prompt & $p_{\mathrm{tool}}$ & Probe~2 $(Low, Med, High)$ \\
\midrule
4 & Upcoming IPO listings & 0.846 & (0.992, 0.005, 0.003) \\
7 & J\&J earnings schedule & 0.632 & (0.996, 0.003, 0.002) \\
10 & Brookdale balance sheet & 0.548 & (0.993, 0.004, 0.004) \\
12 & TTD balance sheet / liquidity & 0.656 & (0.988, 0.012, 0.000) \\
14 & Tesla income statement & 0.881 & (0.997, 0.003, 0.000) \\
15 & Follow-up after Tesla request & 0.202 & (0.438, 0.484, 0.078) \\
\bottomrule
\end{tabularx}
\end{table}
\vspace{-0.8em}

Table~\ref{tab: financial_example_main} summarizes the key steps, abbreviated prompts, and probe outputs.

The trace illustrates the intended runtime behavior. Retrieval-heavy steps have elevated Tool-Need probabilities, ranging from 0.548 to 0.881, while Probe~2 remains almost entirely low-risk, with low-risk probabilities near 0.99 on most retrieval steps. Step 15 is different: it is a follow-up turn, $p_{\mathrm{tool}}$ falls to 0.202, and the risk distribution becomes mixed rather than retrieval-like. Thus the example shows that Tool-Need tracks whether fresh external information is needed, while Tool-Risk is not simply duplicating the tool-need score.

\subsection{Layer and feature analysis}
\hypertarget{sec: results_layers}{}\label{sec: results_layers}

We then examine where the monitoring signal is concentrated and which interpretable SAE features contribute most strongly to the readout. As shown in Table~\ref{tab: layer_concentration}, both models concentrate their tool-decision signal in late transformer layers, with the top Probe~1 features clustering toward the final monitored layers in both backbones.

\begin{table}[H]
\centering
\scriptsize
\setlength{\tabcolsep}{4pt}
\caption{Layer concentration of top-20 Probe-1 features by model.}
\label{tab: layer_concentration}
\begin{minipage}[t]{0.45\linewidth}
\centering
\begin{tabular}{@{}lrrrrrr@{}}
\toprule
\multicolumn{7}{c}{\textbf{GPT-OSS}} \\
Layer & L3 & L7 & L11 & L15 & L19 & \textbf{L23} \\
\midrule
Count & 0 & 0 & 2 & 1 & 4 & \textbf{13} \\
\bottomrule
\end{tabular}
\end{minipage}
\hfill
\begin{minipage}[t]{0.45\linewidth}
\centering
\begin{tabular}{@{}lrrrr@{}}
\toprule
\multicolumn{5}{c}{\textbf{Gemma}} \\
Layer & L16 & L31 & \textbf{L40} & \textbf{L53} \\
\midrule
Count & 0 & 5 & \textbf{7} & \textbf{7} \\
\bottomrule
\end{tabular}
\end{minipage}
\end{table}

This pattern suggests late-stage decision encoding: 13 of GPT-OSS's top-20 Tool-Need features occur in layer 23, while Gemma's top features concentrate in layers 40 and 53. The strongest tool-decision features therefore appear close to the point where the model commits to answering directly or delegating to a tool.

Table~\ref{tab: features_p1} reports representative top SAE features for Probe~1 (Tool-Need), and Table~\ref{tab: features_risk} reports the corresponding representative features for Probe~2 (Tool-Risk).

\begin{table}[H]
\centering
\scriptsize
\setlength{\tabcolsep}{3pt}
\caption{Representative top SAE features for Probe~1 (Tool-Need), shown separately for GPT-OSS and Gemma.}
\label{tab: features_p1}
\begin{minipage}[t]{0.48\linewidth}
\centering
\begin{tabularx}{\linewidth}{@{}p{0.14\linewidth}p{0.20\linewidth}X@{}}
\toprule
\multicolumn{3}{c}{\textbf{GPT-OSS} (top 5)} \\
Layer & Feature & Label \\
\midrule
23 & 79,265 & mathematical expressions, numbers \\
23 & 90,074 & numbers and numerical data \\
23 & 106,054 & legal and formal language \\
23 & 106,420 & numerical values, sequences \\
23 & 38,964 & coordinates, measurements \\
\bottomrule
\end{tabularx}
\end{minipage}
\hfill
\begin{minipage}[t]{0.48\linewidth}
\centering
\begin{tabularx}{\linewidth}{@{}p{0.14\linewidth}p{0.20\linewidth}X@{}}
\toprule
\multicolumn{3}{c}{\textbf{Gemma} (top 5)} \\
Layer & Feature & Label \\
\midrule
53 & 1,694 & punctuation, numbers, list sequences \\
40 & 204 & numerical data: quantities, years \\
40 & 1,084 & high-freq common words, punctuation \\
53 & 2,322 & professional training, evaluative terms \\
40 & 166 & nouns: projects, systems, structures \\
\bottomrule
\end{tabularx}
\end{minipage}
\end{table}

Next, we inspect which internal SAE features correspond to plausible tool-need and tool-risk. These feature tables matter because they show that the probes are not only predictive but also inspectable at the feature level. Table~\ref{tab: features_p1} highlights the numerical and formal-language features associated with tool-call decisions.

Table~\ref{tab: features_risk} shows that the strongest Probe~2 features emphasize authentication, account, and credential-related concepts rather than tool names alone. This suggests that Probe~2 is reading cues about action consequence from the surrounding context, rather than relying only on static tool names. These feature-level views are the main practical reason to retain the SAE-based monitor despite the stronger predictive performance of raw residual probes reported in Appendix~\ref{sec: appendix_raw_probe_baseline}. The raw probe is a stronger classifier in the current run, but it does not directly provide the same layer-level and sparse-feature evidence needed for audit, escalation, or root-cause analysis in high-stakes workflows.

\begin{table}[H]
\centering
\scriptsize
\setlength{\tabcolsep}{3pt}
\caption{Representative top SAE features for Probe~2 (Tool-Risk), shown separately for GPT-OSS and Gemma.}
\label{tab: features_risk}
\begin{minipage}[t]{0.48\linewidth}
\centering
\begin{tabularx}{\linewidth}{@{}p{0.14\linewidth}p{0.20\linewidth}X@{}}
\toprule
\multicolumn{3}{c}{\textbf{GPT-OSS}} \\
Layer & Feature & Label \\
\midrule
23 & 63,701 & password generation / programming \\
23 & 58,277 & notable figures, formal declarations \\
19 & 80,633 & motivation / overcoming challenges \\
23 & 106,054 & legal / formal policy language \\
23 & 38,964 & coordinates, measurements, data \\
\bottomrule
\end{tabularx}
\end{minipage}
\hfill
\begin{minipage}[t]{0.48\linewidth}
\centering
\begin{tabularx}{\linewidth}{@{}p{0.14\linewidth}p{0.20\linewidth}X@{}}
\toprule
\multicolumn{3}{c}{\textbf{Gemma}} \\
Layer & Feature & Label \\
\midrule
53 & 6,032 & usernames, passwords, authentication \\
53 & 10,969 & password creation, security management \\
53 & 3,246 & logging in, accounts, authentication \\
53 & 21,923 & password formats and examples \\
40 & 10,969 & password / account security \\
\bottomrule
\end{tabularx}
\end{minipage}
\end{table}

\subsection{Feature Ablation and Functional Importance}
\hypertarget{sec: results_ablation}{}

Feature ablation assesses whether the selected SAE features materially affect the monitor's output, rather than merely correlating with it. Table~\ref{tab: ablation} reports two effects: ``Flips'' counts how often removing selected SAE features changes the binary Tool-Need decision, and mean $|\Delta p|$ measures the average change in Tool-Need probability.

\begin{table}[H]
\centering
\scriptsize
\setlength{\tabcolsep}{4pt}
\caption{Tool-Need Probe ablation results (10 held-out Nemotron steps per model). $\Delta p$ = mean $|\Delta p_{\mathrm{tool}}|$; Flip = binary prediction flips out of 10.}
\label{tab: ablation}
\begin{minipage}[t]{0.45\linewidth}
\centering
\begin{tabular}{@{}lccc@{}}
\toprule
\multicolumn{4}{c}{\textbf{GPT-OSS}} \\
Set & \# Latents & Flips & Mean $|\Delta p|$ \\
\midrule
Top-5 & 5 & 3/10 & 0.236 \\
Top-10 & 10 & \textbf{4/10} & \textbf{0.431} \\
Top-20 & 20 & 4/10 & 0.384 \\
\bottomrule
\end{tabular}
\end{minipage}
\hfill
\begin{minipage}[t]{0.45\linewidth}
\centering
\begin{tabular}{@{}lccc@{}}
\toprule
\multicolumn{4}{c}{\textbf{Gemma}} \\
Set & \# Latents & Flips & Mean $|\Delta p|$ \\
\midrule
Top-50 & 50 & 0/10 & 0.031 \\
Top-100 & 100 & 0/10 & 0.058 \\
Top-200 & 200 & \textbf{1/10} & \textbf{0.146} \\
\bottomrule
\end{tabular}
\end{minipage}
\end{table}

For GPT-OSS, removing the top 10 features flips 4 of 10 decisions and changes $p_{\mathrm{tool}}$ by 0.431 on average, showing that a compact feature set carries much of the Tool-Need signal. Gemma is more distributed: even the top 200 features flip only 1 of 10 decisions, though the probability shift increases to 0.146. This suggests the GPT-OSS probe relies on a more concentrated signal, while Gemma spreads the signal across a broader sparse feature set. Random-feature ablation produces negligible effects, supporting the claim that the identified features are specific components of the signal the probe reads out.

\section{Runtime Monitoring and Cross-Dataset Evaluation}
\hypertarget{sec: runtime_section}{}\label{sec: runtime_section}

\noindent The monitoring layer that turns probe signals into runtime actions was introduced in Section~\ref{sec: methodology} (Figure~\ref{fig: framework_overview}); here we evaluate it under same-distribution replay and zero-shot transfer.

\subsection{Training and transfer setup}
\hypertarget{sec: runtime_setup}{}

Probes are trained on \textbf{Nemotron} data only. Held-out \textbf{Nemotron replay} provides the same-distribution runtime check for GPT-OSS~20B, while \textbf{BFCL} is used as a strict zero-shot transfer benchmark with no retraining or threshold tuning (\rcite{bfcl2025}{Patil et al., 2025}). This section therefore separates familiar-distribution replay from cross-benchmark transfer.

\vspace{-0.45em}
\subsection{In-Distribution Runtime Evaluation}
\hypertarget{sec: runtime_indomain}{}

We first evaluate the monitor on held-out Nemotron replay using GPT-OSS~20B, the same model family used for activation extraction. Table~\ref{tab: nem_episode_rt} separates delegation errors from tool-formatting errors.
\vspace{-0.8em}

\begin{table}[H]
\centering
\scriptsize
\setlength{\tabcolsep}{6pt}
\setlength{\abovecaptionskip}{1pt}
\setlength{\belowcaptionskip}{1pt}
\caption{Held-out Nemotron replay runtime profile (760 episodes, GPT-OSS).}
\label{tab: nem_episode_rt}
\begin{tabular}{@{}l l@{}}
\toprule
\textbf{Metric} & \textbf{Value} \\
\midrule
Step accuracy & 78.6\% \\
Missed-tool-call rate & 34.2\% of tool-required steps \\
Unnecessary-call rate & 7.7\% of no-tool steps \\
Tool-naming accuracy (given call) & 90.8\% \\
Missed-tool cases flagged by probe & ${\sim}$75.5\% (${\sim}$258/343) \\
\bottomrule
\end{tabular}
\end{table}
\vspace{-1.25em}
The model misses 34.2\% of tool-required steps, but once it decides to call a tool, tool naming is much stronger at 90.8\%. The probe flags about 75.5\% of missed-tool cases, suggesting that Probe~1 is most useful as a pre-action omission monitor rather than a tool-name checker. Operationally, these missed- and unnecessary-call rates are the \emph{Action}-box outcomes of Figure~\ref{fig: agent_anatomy} measured live—caught before execution rather than read from the logs afterward.

\subsection{Out-of-Distribution Evaluation on BFCL}
\hypertarget{sec: runtime_bfcl}{}

BFCL is used here as a strict zero-shot transfer benchmark: the probes are trained on Nemotron only and then evaluated on a different benchmark format without retraining, calibration, or threshold tuning. Instead, BFCL multi-turn episodes are mapped into the same step-level format used for Nemotron: cumulative transcript becomes \texttt{context}, gold call annotations determine \texttt{tool\_needed}, and BFCL tools are heuristically projected into the Nemotron low/medium/high risk-tier scheme for Probe~2. This preserves the same pre-action setup under a different benchmark distribution.
\vspace{-0.7em}

\begin{table}[H]
\centering
\scriptsize
\setlength{\tabcolsep}{6pt}
\setlength{\abovecaptionskip}{1pt}
\setlength{\belowcaptionskip}{1pt}
\caption{BFCL zero-shot transfer summary.}
\label{tab: bfcl_transfer_summary}
\begin{tabular}{@{}lcc@{}}
\toprule
\textbf{Metric} & \textbf{GPT-OSS 20B} & \textbf{Gemma 3 27B} \\
\midrule
\multicolumn{3}{@{}l}{\textit{Runtime performance}} \\
Expected--Actual agreement & 83.6\% & 87.6\% \\
Missed-tool rate (tool steps) & 10.8\% & 0.2\% \\
Unnecessary-call rate (no-tool steps) & 56.3\% & 98.5\% \\
Episode success (all steps correct) & 49.2\% & 50.0\% \\
Mean first-failure step & 1.69 turns & 1.17 turns \\
\midrule
\multicolumn{3}{@{}l}{\textit{Probe quality}} \\
Probe-1 agreement with gold (E vs I) & 87.7\% & 77.7\% \\
\bottomrule
\end{tabular}
\end{table}
\vspace{-1.0em}
Table~\ref{tab: bfcl_transfer_summary} summarizes the transfer results, while Appendix~\ref{sec: appendix_qualitative} includes Table~\ref{tab: nemotron_bfcl_format_example}, an illustrative formatting contrast between the two benchmark styles.

The main transfer result is that Probe~1 remains aligned with gold tool need on BFCL: expected--internal agreement is 87.7\% for GPT-OSS and 77.7\% for Gemma. However, the runtime profile differs by model. GPT-OSS has a 10.8\% missed-tool rate but also a 56.3\% unnecessary-call rate, while Gemma nearly eliminates missed tools at 0.2\% but over-triggers on 98.5\% of no-tool steps. Thus BFCL shows useful omission sensitivity under distribution shift, but also confirms that thresholds require recalibration on a new tool distribution.

A representation comparison on the same BFCL zero-shot slice is reported in Appendix~\ref{sec: appendix_raw_probe_baseline}. Unlike the held-out Nemotron comparison, the out-of-distribution results do not show a uniform probe advantage: relative performance varies across backbone and target. This suggests that both dense residual and SAE representations retain useful pre-action tool-decision signal under distribution shift, while their relative predictive performance is model and task-dependent.

Probe~2 transfers less cleanly than Tool-Need, which is expected because risk labels depend more heavily on how tool categories are mapped across datasets. Failures are early in both models: mean first-failure occurs within the first one or two turns, including \textbf{1.17} turns for Gemma on the merged BFCL slice. This suggests a short but practically relevant intervention window for pre-execution monitoring. Overall, BFCL should be read as a transfer stress test: Probe~1 transfers best as an omission-auditing signal at the tool-call boundary, while Probe~2 remains useful for separating lower- from higher-risk actions once a tool call is in play.

\section{Discussion and Limitations}
\hypertarget{sec: discussion}{}\label{sec: discussion}

The main value of this framework is that it provides a \emph{pre-action} view of tool decisions. External monitoring, logs, and benchmark scores remain useful, but they mostly explain behavior after the model has already acted. Our probes instead read internal state at the decision boundary itself, before execution. This is the tools-and-action boundary of Figure~\ref{fig: agent_anatomy}: the pre-action layer that logs and output scoring miss. This matters because many agent failures are trajectory-shaping: an early missed tool call or unnecessary tool use can change the context seen by every later step and produce cascades that output-only monitoring cannot easily disentangle (\rcite{cemri2025}{Cemri et al., 2025}).

A second advantage is generality. Because the monitor operates on internal representations rather than tool-specific output patterns, the same probe framework can apply across multiple tools and repeated decision points. Probe~1 asks whether a tool call is needed at all, while Probe~2 asks whether the next tool action appears more consequential; feature tables and ablations show that these signals are both predictive and localized to late sparse features.

This is where SAE-based monitoring adds a practical advantage over output-only observability. The framework does not simply emit a scalar warning; it identifies where the signal is concentrated, which sparse features are most associated with the decision, and whether suppressing those features changes the probe output.

This does not mean that SAE+probe is always more accurate than every simpler baseline. Its main advantage is that it changes the type of evidence available to the operator. A raw residual-stream probe may answer whether a decision signal is linearly readable, but it usually cannot explain which sparse internal features contributed to the warning. Output-based monitors and tool-call probabilities are even further removed from the model's internal decision state. SAE+probe is therefore best understood as an observability trade-off: it introduces additional activation-level complexity, but provides feature-level evidence that is valuable when monitoring decisions must be reviewed, audited, or debugged after failure.

The runtime results suggest that the framework is most useful as an oversight layer at high-value decision points. Held-out Nemotron replay shows that the main bottleneck is deciding to delegate at all, not naming a tool once delegation has begun. BFCL then serves as a transfer stress test: Probe~1 transfers best as an omission-auditing signal, while Probe~2 remains useful for risk-tiering once a tool call is underway. We therefore interpret the Nemotron--BFCL gap as reflecting both task transfer and added benchmark-specific instruction-following pressure, not simply loss of the internal signal itself.

\paragraph{Limitations.}
Tool-Need is the stronger and more stable probe. Appendix~\ref{sec: appendix_raw_probe_baseline} adds a preliminary raw residual-stream probe baseline. On held-out Nemotron, raw probes outperform SAE-feature probes on predictive metrics. This reinforces an important limitation: the present paper should not be read as claiming that SAEs are necessary for maximum classification accuracy. The claim is narrower: SAE-based probes provide an inspectable pre-action monitoring layer with sparse features, layer localization, and feature-level evidence. Future work can broaden this comparison across additional models, layers, datasets, and agent environments. Tool-Risk is more sensitive because it depends on a heuristic action-risk scheme rather than a universal standard for tool consequence. Tool-Risk transfer is not only a benchmark shift: different tools, prompting protocols, scoring rules, and agent environments may encode tool need and risk differently. The paper also studies two open-weight backbones and a selected set of layers, so broader portability across architectures, scales, layer choices, and post-training recipes remains an open question. We also deliberately scope to the tool-decision boundary; the \emph{memory} and \emph{planning} dynamics shown faded in Figure~\ref{fig: agent_anatomy} are out of scope here and are a natural direction for future work, since failures there can also propagate through a trajectory. Finally, feature identities may drift with checkpoint choice and SAE recipe, even when late-layer concentration appears robust. Additional examples and per-step tables appear in Appendix~\ref{sec: appendix_qualitative}.

\section{Conclusion}
\hypertarget{sec:conclusion}{}

Tool decisions leave readable traces inside the model \emph{before} external execution—at the tools-and-action boundary of Figure~\ref{fig: agent_anatomy}. On held-out Nemotron trajectories, linear probes on SAE-decomposed activations recover Tool-Need and Tool-Risk signals for both \textbf{GPT-OSS 20B} and \textbf{Gemma 3 27B IT}, with late-layer concentration and sparse feature sets that affect the monitoring readout under ablation. This provides evidence that the monitor is reading a structured internal signal, while stronger claims about causal control of the model's actual tool-call behavior require future generation-time interventions.

The same monitoring recipe transfers across backbones under distinct SAE variants and layer selections and remains informative under live replay and BFCL out-of-distribution evaluation, with Tool-Need acting most reliably as an \textbf{omission auditor} and Tool-Risk as a \textbf{risk-oriented} layer that requires extra care when the risk scheme shifts across tool namespaces. Selective activation capture at decision boundaries therefore offers a practical complement to external benchmarks and execution logs for safer, more controllable agent deployment.

The broader contribution is to show that mechanistic interpretability can become \textbf{operationally useful} for agent systems: extending our earlier \textit{Beyond the Black Box} study of LLM interpretability (\rcite{tatsat2025}{Tatsat \& Shater, 2025}) into agent settings, this paper shows how internal monitoring can move beyond explaining behavior after failure to helping monitor tool decisions before action in realistic high-stakes workflows.

\clearpage
\section*{References}
\small

\noindent \hypertarget{ref: alain2017}{}Alain, G., \& Bengio, Y. (2017). Understanding Intermediate Layers Using Linear Classifier Probes. \textit{arXiv preprint arXiv:1610.01644}. \href{https://arxiv.org/abs/1610.01644}{https://arxiv.org/abs/1610.01644}

\noindent \hypertarget{ref: bricken2023}{}Bricken, T., et al. (2023). Towards Monosemanticity: Decomposing Language Models with Dictionary Learning. \textit{Transformer Circuits Thread}. \href{https://transformer-circuits.pub/2023/monosemantic-features/}{https://transformer-circuits.pub/2023/monosemantic-features/}

\noindent \hypertarget{ref: cemri2025}{}Cemri, M., et al. (2025). Why Do Multi-Agent LLM Systems Fail? \textit{arXiv preprint arXiv:2503.13657}. \href{https://arxiv.org/abs/2503.13657}{https://arxiv.org/abs/2503.13657}

\noindent \hypertarget{ref: nemotron2026}{}Chandiramani, A., et al. (2026). Nemotron 3 Super: Open, Efficient Mixture-of-Experts Hybrid Mamba-Transformer Model for Agentic Reasoning. \textit{arXiv preprint arXiv:2604.12374}. \href{https://arxiv.org/abs/2604.12374}{https://arxiv.org/abs/2604.12374}

\noindent \hypertarget{ref: cho2025}{}Cho, S., Wu, Z., \& Koshiyama, A. (2025). CorrSteer: Steering Improves Task Performance and Safety in LLMs through Correlation-based Sparse Autoencoder Feature Selection. \textit{arXiv preprint arXiv:2508.12535}. \href{https://arxiv.org/abs/2508.12535}{https://arxiv.org/abs/2508.12535}

\noindent \hypertarget{ref: halluc2026}{}Healy, K., et al. (2026). Internal Representations as Indicators of Hallucinations in Agent Tool Selection. \textit{arXiv preprint arXiv:2601.05214}. \href{https://arxiv.org/abs/2601.05214}{https://arxiv.org/abs/2601.05214}

\noindent \hypertarget{ref: meco2025}{}Li, W., et al. (2025). Adaptive Tool Use in Large Language Models with Meta-Cognition Trigger. \textit{arXiv preprint arXiv:2502.12961}. \href{https://arxiv.org/abs/2502.12961}{https://arxiv.org/abs/2502.12961}

\noindent \hypertarget{ref: liu2024}{}Liu, W., et al. (2024). ToolACE: Winning the Points of LLM Function Calling. \textit{arXiv preprint arXiv:2409.00920}. \href{https://arxiv.org/abs/2409.00920}{https://arxiv.org/abs/2409.00920}

\noindent \hypertarget{ref: mckenzie2025}{}McKenzie, A., et al. (2025). Detecting High-Stakes Interactions with Activation Probes. \textit{arXiv preprint arXiv:2506.10805}. \href{https://arxiv.org/abs/2506.10805}{https://arxiv.org/abs/2506.10805}

\noindent \hypertarget{ref: bfcl2025}{}Patil, S. G., et al. (2025). The Berkeley Function Calling Leaderboard (BFCL): From Tool Use to Agentic Evaluation of Large Language Models. \textit{Proceedings of the Forty-Second International Conference on Machine Learning}. \href{https://openreview.net/forum?id=2GmDdhBdDk}{https://openreview.net/forum?id=2GmDdhBdDk}

\noindent \hypertarget{ref: qin2023}{}Qin, Y., et al. (2023). ToolLLM: Facilitating Large Language Models to Master 16000+ Real-World APIs. \textit{arXiv preprint arXiv:2307.16789}. \href{https://arxiv.org/abs/2307.16789}{https://arxiv.org/abs/2307.16789}

\noindent \hypertarget{ref: rai2024}{}Rai, D., et al. (2024). A Practical Review of Mechanistic Interpretability for Transformer-Based Language Models. \textit{arXiv preprint arXiv:2407.02646}. \href{https://arxiv.org/abs/2407.02646}{https://arxiv.org/abs/2407.02646}

\noindent \hypertarget{ref: schick2023}{}Schick, T., et al. (2023). Toolformer: Language Models Can Teach Themselves to Use Tools. \textit{Advances in Neural Information Processing Systems, 36}. \href{https://arxiv.org/abs/2302.04761}{https://arxiv.org/abs/2302.04761}

\noindent \hypertarget{ref: tatsat2025}{}Tatsat, H., \& Shater, A. (2025). Beyond the Black Box: Interpretability of LLMs in Finance. \textit{arXiv preprint arXiv:2505.24650}. \href{https://arxiv.org/abs/2505.24650}{https://arxiv.org/abs/2505.24650}

\noindent \hypertarget{ref: wang2025hammer}{}Wang, J., et al. (2025). HammerBench: Fine-Grained Function-Calling Evaluation in Real Mobile Device Scenarios. \textit{arXiv preprint arXiv:2412.16516}. \href{https://arxiv.org/abs/2412.16516}{https://arxiv.org/abs/2412.16516}

\noindent \hypertarget{ref: epistemic2025}{}Wang, H., et al. (2025). Position: Agent Should Invoke External Tools ONLY When Epistemically Necessary. \textit{arXiv preprint arXiv:2506.00886}. \href{https://arxiv.org/abs/2506.00886}{https://arxiv.org/abs/2506.00886}

\noindent \hypertarget{ref: weng2023}{}Weng, L. (2023). LLM Powered Autonomous Agents. \textit{Lil'Log}. \href{https://lilianweng.github.io/posts/2023-06-23-agent/}{https://lilianweng.github.io/posts/2023-06-23-agent/}

\normalsize

\clearpage
\appendix
\onecolumn
\section*{Appendix}
\addcontentsline{toc}{section}{Appendix}
\vspace{-0.4em}

\section{Raw Residual Probe Baseline}
\hypertarget{sec: appendix_raw_probe_baseline}{}
\label{sec: appendix_raw_probe_baseline}

This appendix reports a preliminary baseline using linear probes trained directly on pooled raw residual-stream activations, without SAE encoding. The purpose of this comparison is to separate two questions: whether tool-decision signals are linearly readable from model activations, and whether SAE features are needed for inspectable monitoring. The raw residual probe is expected to be a strong predictive baseline because it has access to the dense activation vector before sparsification.

Table~\ref{tab: appendix_raw_sae_metrics} compares raw residual probes with the SAE-feature probes used in the main paper. On held-out Nemotron, raw residual probes outperform SAE-feature probes on every reported predictive metric for both GPT-OSS 20B and Gemma 3 27B IT. We therefore do not claim that SAE+probe is the accuracy-maximizing approach. Instead, we use SAE+probe in the main framework because it provides feature-level interpretability, layer localization, and a more inspectable monitoring signal. This distinction is especially important for high-stakes domains such as finance, where a monitoring system must support audit and root-cause analysis rather than only maximize classification accuracy.

The SAE columns in Table~\ref{tab: appendix_raw_sae_metrics} reproduce the held-out SAE-feature results reported in the main paper, while the RAW columns report the new residual-stream probe baseline. The table is intended to make the paper's trade-off explicit: raw residual probes are stronger predictors in this run, while SAE-feature probes are retained because they provide inspectable features and layer-level evidence for monitoring.

\begin{table}[H]
\centering
\scriptsize
\setlength{\tabcolsep}{3pt}
\caption{Preliminary comparison of raw residual-stream probes and SAE-feature probes. Values are reported as RAW / SAE.}
\label{tab: appendix_raw_sae_metrics}
\begin{tabularx}{\linewidth}{@{}>{\raggedright\arraybackslash}p{0.42\linewidth}>{\centering\arraybackslash}p{0.25\linewidth}>{\centering\arraybackslash}p{0.25\linewidth}@{}}
\toprule
\textbf{Metric} & \textbf{Gemma 3 27B IT} & \textbf{GPT-OSS 20B} \\
\midrule
Tool-Need accuracy & \textbf{81.0\%} / 71.4\% & \textbf{81.8\%} / 75.3\% \\
Tool-Need F1 (macro) & \textbf{0.810} / 0.714 & \textbf{0.817} / 0.753 \\
Tool-Risk accuracy (tool rows) & \textbf{96.3\%} / 88.5\% & \textbf{94.0\%} / 90.3\% \\
Tool-Risk macro-F1 & \textbf{0.855} / 0.668 & \textbf{0.799} / 0.699 \\
\bottomrule
\end{tabularx}
\end{table}

\subsection{Zero-shot BFCL comparison}

We additionally compare RAW residual and SAE-feature probes under zero-shot transfer to BFCL. Unlike the held-out Nemotron results, this setting does not show a uniform advantage for either representation. RAW is higher for Gemma Tool-Need agreement and for Tool-Risk agreement in both backbones, while SAE is higher for GPT-OSS Tool-Need agreement. The result is therefore best read as a supplementary robustness check rather than a definitive ranking of representations.

This mixed pattern is consistent with both representations retaining transferable pre-action tool-decision information under distribution shift. Dense residual activations preserve more information before sparsification, whereas SAE features provide a compressed and interpretable representation; under a changed benchmark distribution, neither property guarantees a uniform advantage. Because this evaluation uses a limited BFCL slice, we do not make a broad generalization claim from these differences alone.

\begin{table}[H]
\centering
\scriptsize
\setlength{\tabcolsep}{3pt}
\caption{RAW versus SAE zero-shot BFCL agreement on the same evaluated BFCL decision rows. Values are reported as RAW / SAE; higher is better.}
\label{tab: appendix_raw_sae_bfcl}
\begin{tabularx}{\linewidth}{@{}>{\raggedright\arraybackslash}p{0.42\linewidth}>{\centering\arraybackslash}p{0.25\linewidth}>{\centering\arraybackslash}p{0.25\linewidth}@{}}
\toprule
\textbf{Metric} & \textbf{Gemma 3 27B IT} & \textbf{GPT-OSS 20B} \\
\midrule
Tool-Need agreement with gold & 81.9\% / 77.7\% & 84.7\% / 87.6\% \\
Tool-Risk agreement with gold & 52.9\% / 43.7\% & 52.2\% / 46.6\% \\
\bottomrule
\end{tabularx}
\end{table}

Table~\ref{tab: appendix_gptoss_raw_sae_confusion_toolneed} shows the GPT-OSS Tool-Need confusion matrices for the raw residual and SAE-feature probes. The raw residual probe reduces both missed tool predictions and false tool alerts relative to the SAE-feature probe in this run.

\begin{table}[H]
\centering
\scriptsize
\caption{GPT-OSS 20B Tool-Need confusion matrices for RAW residual and SAE-feature probes. Rows are true labels and columns are predicted labels.}
\label{tab: appendix_gptoss_raw_sae_confusion_toolneed}
\begin{tabular}{lcc}
\toprule
\multicolumn{3}{c}{\textbf{RAW residual probe} (1,935 rows)} \\
 & Pred: no\_tool & Pred: tool \\
\midrule
True: no\_tool & 763 & 170 \\
True: tool & 183 & 819 \\
\midrule
\multicolumn{3}{c}{\textbf{SAE-feature probe} (1,993 rows)} \\
 & Pred: no\_tool & Pred: tool \\
\midrule
True: no\_tool & 741 & 225 \\
True: tool & 267 & 760 \\
\bottomrule
\end{tabular}
\end{table}

Table~\ref{tab: appendix_gptoss_raw_sae_confusion_risk} shows the GPT-OSS Tool-Risk confusion matrices. The raw residual probe improves classification of the low- and high-risk classes and also improves the medium-risk class in this run, although the medium class remains much smaller than the low-risk class.

\begin{table}[H]
\centering
\scriptsize
\caption{GPT-OSS 20B Tool-Risk confusion matrices for RAW residual and SAE-feature probes. Rows are true labels and columns are predicted labels.}
\label{tab: appendix_gptoss_raw_sae_confusion_risk}
\begin{tabular}{lccc}
\toprule
\multicolumn{4}{c}{\textbf{RAW residual probe} (964 tool rows)} \\
 & Pred: low & Pred: medium & Pred: high \\
\midrule
True: low & 812 & 11 & 7 \\
True: medium & 20 & 30 & 1 \\
True: high & 11 & 8 & 64 \\
\midrule
\multicolumn{4}{c}{\textbf{SAE-feature probe} (987 tool rows)} \\
 & Pred: low & Pred: medium & Pred: high \\
\midrule
True: low & 818 & 40 & 18 \\
True: medium & 16 & 24 & 2 \\
True: high & 17 & 3 & 49 \\
\bottomrule
\end{tabular}
\end{table}

Taken together, the in-distribution and zero-shot comparisons support a trade-off rather than a simple dominance claim. RAW residual probes are stronger on the held-out Nemotron comparison, while the BFCL zero-shot results are mixed across model backbones and targets. SAE+probe is therefore justified in this paper primarily as an interpretability and observability choice, not as a universally superior classification representation.

\section{FEATURE LABELING METHODOLOGY}
\hypertarget{sec: appendix_autointerp}{}
\label{sec: appendix_autointerp}

Table~\ref{tab: autointerp_method} summarizes the paper's \emph{feature labeling methodology} for representative SAE features. We use it only to provide a small number of compact feature labels that help the reader interpret the probe tables.

For each selected feature, we retrieve top-activating examples, convert them into a short evidence summary, and use that evidence to produce a concise candidate label. When available, we also use "Neuronpedia" as an external automated-interpretability browser for inspecting SAE features and activation examples; its descriptions are treated only as a cross-check rather than as ground truth. The resulting labels are therefore best understood as brief interpretability anchors, not as definitive names for model circuits.

\begin{table}[H]
\centering
\small
\caption{Technical summary of the feature labeling methodology used for representative SAE features.}
\label{tab: autointerp_method}
\begin{tabularx}{\linewidth}{@{}p{0.18\linewidth}p{0.22\linewidth}X@{}}
\toprule
\textbf{Stage} & \textbf{Object} & \textbf{Technical role} \\
\midrule
Feature specification & model ID + SAE source + feature index & Defines the exact latent to be labeled. \\
Activation retrieval & top-activating examples & Collects the strongest evidence for that feature from Neuronpedia, local inference, or precomputed activations. \\
Evidence formatting & token/value pairs or top text spans & Converts raw activation evidence into a compact explanation prompt. \\
Label generation & local or API-based labeling model & Produces a concise candidate description for the selected feature. \\
Cross-checking & candidate label + retrieved evidence & Checks that the label matches the strongest activation contexts. \\
Output & short feature label & Reports a compact interpretability anchor for representative features in the paper. \\
\bottomrule
\end{tabularx}
\end{table}

\section{NEMOTRON RISK-TIER SCHEME}
\hypertarget{sec: appendix_risk_scheme}{}\label{sec: appendix_risk_scheme}

This appendix summarizes the keyword groups used to instantiate the Nemotron risk-tier scheme referenced in Section~\ref{sec: dataset}. The scheme is heuristic and is intended to capture the operational risk of the next tool action rather than its domain label. Table~\ref{tab: appendix_risk_scheme_summary} lists the tier definitions and representative keyword groups.

\begin{table}[H]
\centering
\small
\caption{Keyword groups used in the Nemotron risk-tier scheme. Representative tools are illustrative rather than exhaustive.}
\label{tab: appendix_risk_scheme_summary}
\renewcommand{\arraystretch}{1.15}
\begin{tabularx}{\linewidth}{@{}p{0.12\linewidth}p{0.13\linewidth}>{\raggedright\arraybackslash}p{0.24\linewidth}>{\raggedright\arraybackslash}X@{}}
\toprule
\textbf{Tier} & \textbf{Distinct tools} & \textbf{Operational meaning} & \textbf{Representative keyword groups / tools} \\
\midrule
Low & 920 & Read-only retrieval, lookup, and search operations with no persistent state change & Search, retrieval, lookup, financial data access, read-only inspection \\
\addlinespace[3pt]
Medium & 27 & Bounded write or creation operations with contained scope & write\_file, text\_editor, create\_desktop\_txt\_file, run-code, generateImageUrl, image\_generation, generateImage, create\_room, create\_subdirectory, save\_as\_pdf \\
\addlinespace[3pt]
High & 21 & Authentication, outbound communication, or dangerous execution actions with external consequence or irreversibility & getusertoken, registeruser, modifypassword, forgotpassword, deleteaccount, sendemail, sendim, sendmessage, shell-exec, execute\_command, python\_exec, exec, execute\_bash\_code, run\_zapier\_NLA\_action \\
\bottomrule
\end{tabularx}
\end{table}

\section{ADDITIONAL FINANCIAL QUALITATIVE EXAMPLES}
\hypertarget{sec: appendix_qualitative}{}\label{sec: appendix_qualitative}

This appendix collects the supplementary qualitative traces that support the worked example in Section~\ref{sec: results_finance}. Figure~\ref{fig: appendix_finance_probe1} covers the multi-ticker fundamentals trace, Figures~\ref{fig: appendix_dca_probe1} and~\ref{fig: appendix_dca_probe2} cover the Bitcoin DCA trace, Table~\ref{tab: nemotron_bfcl_format_example} provides the illustrative Nemotron--BFCL formatting contrast referenced in Section~\ref{sec: runtime_section}, and Tables~\ref{tab: appendix_nemotron_financial_full}--\ref{tab: appendix_bfcl_trading_full} provide the corresponding full per-step values.

\subsection{Illustrative cross-benchmark formatting example}

Table~\ref{tab: nemotron_bfcl_format_example} shows the non-evaluative formatting contrast referenced in Section~\ref{sec: runtime_section}. It is included only to illustrate how similar tool-use logic can appear under different benchmark representations.

\begin{table}[H]
\centering
\scriptsize
\setlength{\tabcolsep}{3pt}
\caption{Illustrative formatting contrast between Nemotron and BFCL. The table is included only to show how similar tool-use logic can appear under different benchmark representations; it is not itself an evaluation result.}
\label{tab: nemotron_bfcl_format_example}
\begin{tabularx}{0.98\linewidth}{@{}p{0.18\linewidth}X@{}}
\toprule
\textbf{Dataset} & \textbf{Illustrative prompt format} \\
\midrule
Nemotron (trajectory-style) & \textbf{User:} ``I have \$450, spent \$150 on groceries and \$50 on utilities. How much remains, and what is 20\% for emergency savings?'' \textbf{Reasoning trace:} compute remaining amount, then compute 20\%. \textbf{Tool calls:} \texttt{subtract(\{a:450,b:150\})} $\rightarrow$ 300; \texttt{subtract(\{a:300,b:50\})} $\rightarrow$ 250. \textbf{Assistant:} ``You have \$250 left; save \$50.'' \\
BFCL (benchmark-style) & \texttt{question}: ``Read finance values and compute remaining cash; then compute 20\% savings.'' \texttt{ground\_truth}: [\texttt{subtract(a=450,b=150)}, \texttt{subtract(a=300,b=50)}, \texttt{echo(content='50',file\_name='savings.txt')}] \\
\bottomrule
\end{tabularx}
\end{table}

\subsection{Multi-ticker fundamentals trace (\texttt{trajectory\_id}~3344)}

\begin{figure}[t]
\centering
\includegraphics[width=\linewidth]{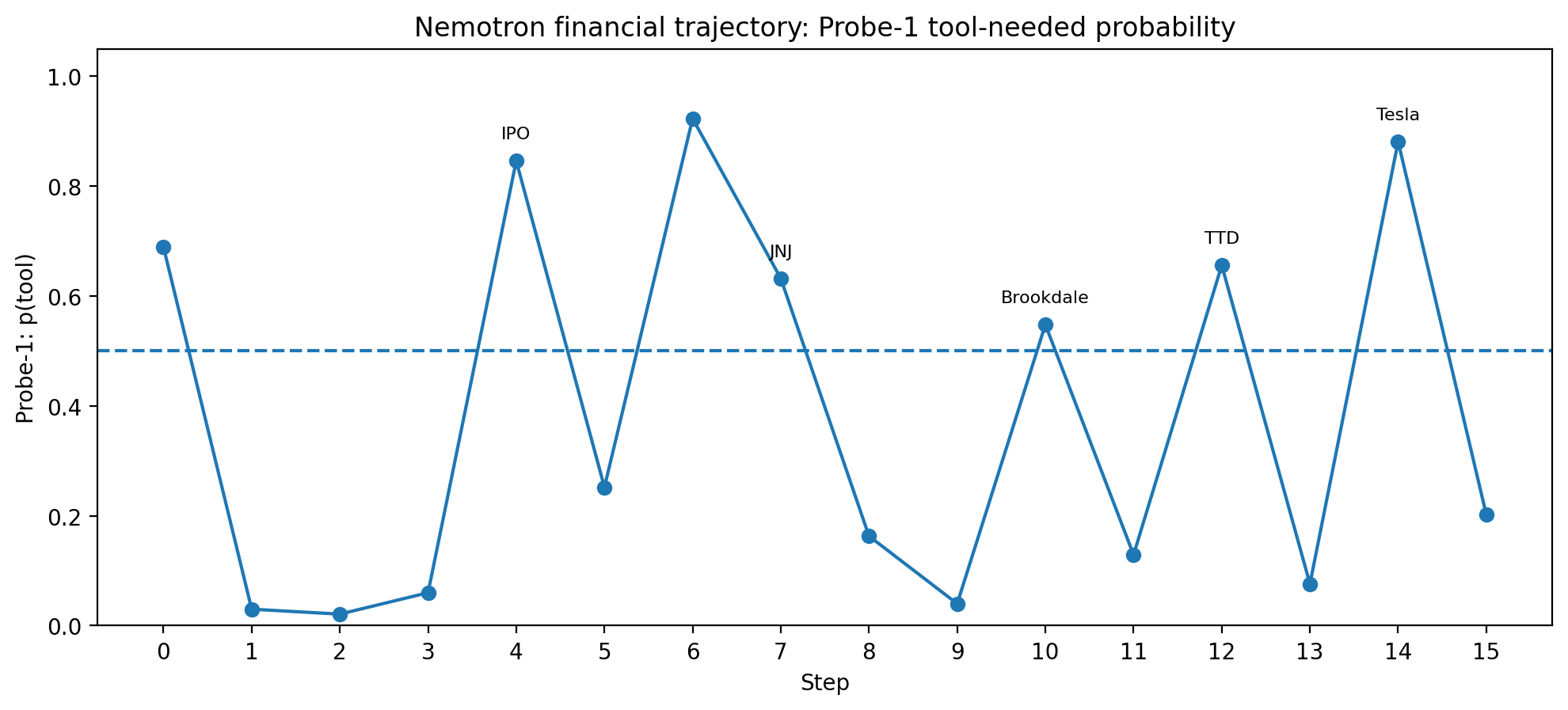}
\caption{Tool-Need Probe (Probe~1) on the multi-ticker fundamentals trajectory. The signal rises on steps that require external financial retrieval and falls on follow-up no-tool steps.}
\label{fig: appendix_finance_probe1}
\end{figure}

Figure~\ref{fig: appendix_finance_probe1} shows the Tool-Need probability curve for the multi-ticker fundamentals trace discussed in Section~\ref{sec: results_finance}; the full step-level values appear in Table~\ref{tab: appendix_nemotron_financial_full}.

\subsection{Bitcoin DCA scenario (Nemotron, \texttt{trajectory\_id}~4592)}

\begin{figure}[t]
\centering
\includegraphics[width=0.92\linewidth]{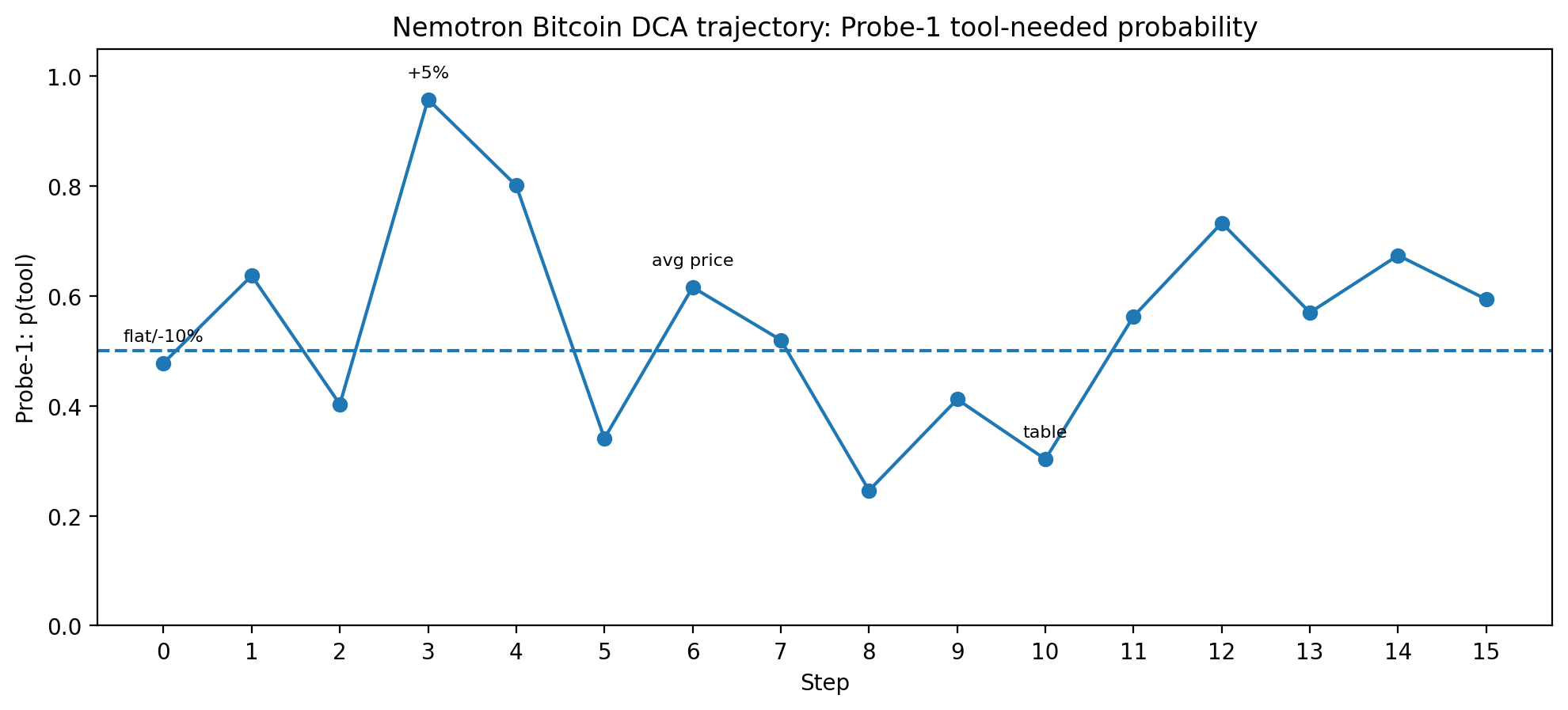}
\caption{Tool-Need Probe (Probe~1) on the Bitcoin DCA trajectory. The signal rises on calculation-heavy steps.}
\label{fig: appendix_dca_probe1}
\end{figure}

\begin{figure}[t]
\centering
\includegraphics[width=0.92\linewidth]{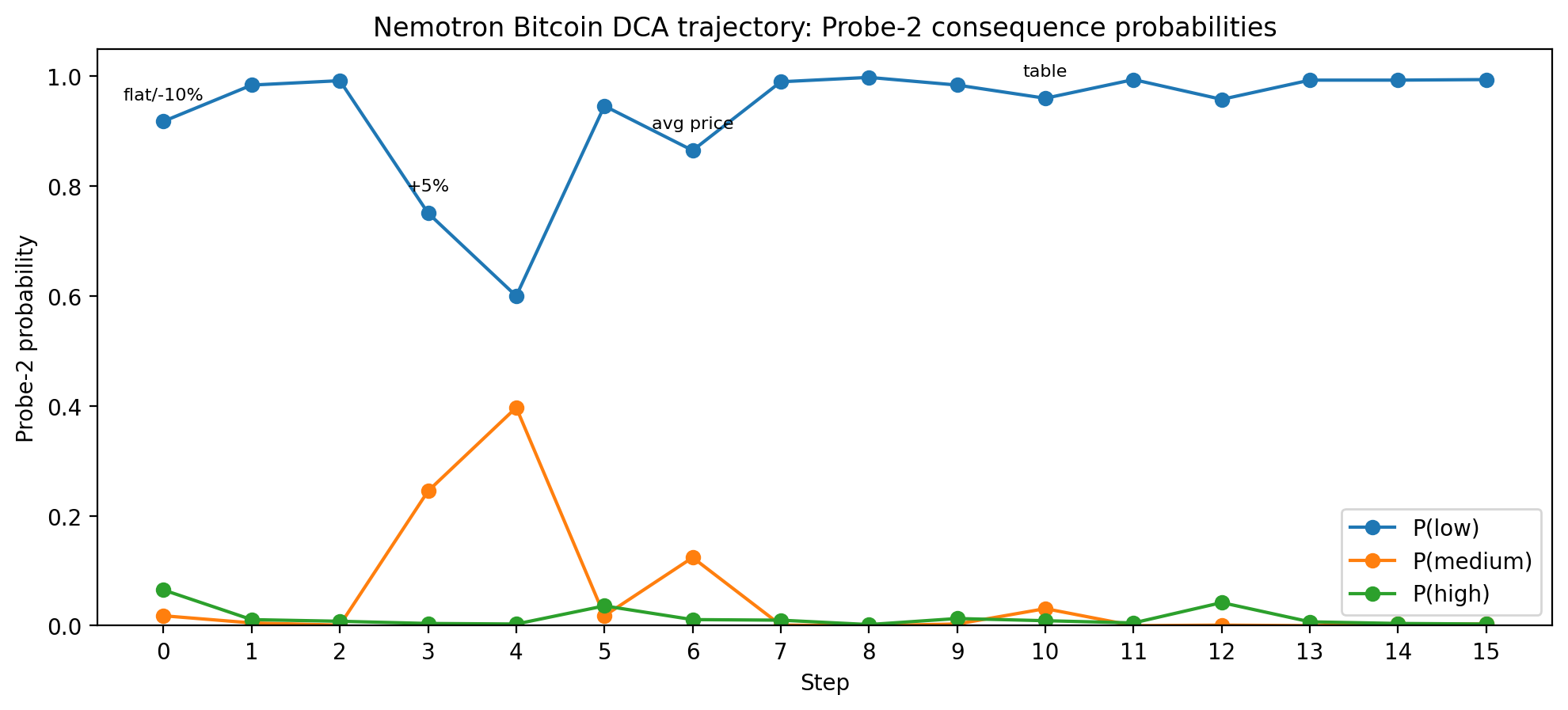}
\caption{Tool-Risk Probe (Probe~2) on the Bitcoin DCA trajectory. Risk remains low for calculator-style actions.}
\label{fig: appendix_dca_probe2}
\end{figure}

Figures~\ref{fig: appendix_dca_probe1} and~\ref{fig: appendix_dca_probe2} show the corresponding Tool-Need and Tool-Risk traces for the Bitcoin DCA scenario; the full step-level values appear in Table~\ref{tab: appendix_nemotron_dca_full}.

\subsection{Full per-step probe tables}

Tables~\ref{tab: appendix_nemotron_financial_full}, \ref{tab: appendix_nemotron_dca_full}, and \ref{tab: appendix_bfcl_trading_full} report the full step-level outputs underlying the qualitative examples and BFCL slice referenced in the main text.

{\footnotesize
\begin{table}[H]
\centering
\caption{Nemotron multi-ticker fundamentals (\texttt{trajectory\_id}~3344): all pivot steps (GPT-OSS probes). ``Phase'' summarizes the latest user intent at each step; tier = Probe-2 argmax.}
\label{tab: appendix_nemotron_financial_full}
\begin{tabularx}{\linewidth}{@{}r>{\raggedright\arraybackslash}Xccccc@{}}
\toprule
Step & Phase / user intent & $p_{\mathrm{tool}}$ & Tier & $p_{\mathrm{low}}$ & $p_{\mathrm{med}}$ & $p_{\mathrm{high}}$ \\
\midrule
0 & Confirm SU + API key after Suncor ask & 0.689 & low & 0.999 & 0.000 & 0.001 \\
1 & Same turn (extended context) & 0.030 & low & 0.886 & 0.019 & 0.094 \\
2 & Emerson EMR: income \& cash flow & 0.021 & low & 0.950 & 0.031 & 0.020 \\
3 & IBM: overview \& ratios & 0.060 & low & 0.990 & 0.007 & 0.003 \\
4 & IPO names / listings & 0.846 & low & 0.992 & 0.005 & 0.003 \\
5 & Same IPO thread & 0.252 & low & 0.985 & 0.001 & 0.015 \\
6 & Pivot: healthcare / senior living & 0.923 & low & 0.936 & 0.049 & 0.016 \\
7 & JNJ: earnings schedule (12 months) & 0.632 & low & 0.996 & 0.003 & 0.002 \\
8 & Same JNJ request & 0.163 & low & 0.939 & 0.033 & 0.028 \\
9 & Same JNJ request & 0.040 & low & 0.881 & 0.001 & 0.118 \\
10 & Brookdale: balance sheet & 0.548 & low & 0.993 & 0.004 & 0.004 \\
11 & Same Brookdale thread & 0.129 & low & 0.988 & 0.003 & 0.009 \\
12 & The Trade Desk (TTD): balance sheet / liquidity & 0.656 & low & 0.988 & 0.012 & 0.000 \\
13 & Same TTD thread & 0.076 & low & 0.868 & 0.006 & 0.126 \\
14 & Tesla: income statement \& revenue trends & 0.881 & low & 0.997 & 0.003 & 0.000 \\
15 & Same Tesla thread & 0.202 & medium & 0.438 & 0.484 & 0.078 \\
\bottomrule
\end{tabularx}
\end{table}

\begin{table}[H]
\centering
\caption{Nemotron Bitcoin DCA scenario (\texttt{trajectory\_id}~4592): all pivot steps (GPT-OSS probes).}
\label{tab: appendix_nemotron_dca_full}
\begin{tabularx}{\linewidth}{@{}r>{\raggedright\arraybackslash}Xccccc@{}}
\toprule
Step & Phase / user intent & $p_{\mathrm{tool}}$ & Tier & $p_{\mathrm{low}}$ & $p_{\mathrm{med}}$ & $p_{\mathrm{high}}$ \\
\midrule
0 & Opening DCA question (flat vs.\ 10\%/mo decline) & 0.478 & low & 0.918 & 0.018 & 0.065 \\
1 & Same opening (more tool traces in context) & 0.637 & low & 0.984 & 0.005 & 0.011 \\
2 & Same opening & 0.403 & low & 0.992 & 0.001 & 0.008 \\
3 & Add +5\%/month price path & 0.958 & low & 0.751 & 0.245 & 0.004 \\
4 & Same follow-up & 0.801 & low & 0.600 & 0.397 & 0.003 \\
5 & Same follow-up & 0.341 & low & 0.946 & 0.018 & 0.036 \\
6 & Average purchase price per BTC, three scenarios & 0.616 & low & 0.865 & 0.124 & 0.011 \\
7 & Same & 0.520 & low & 0.990 & 0.001 & 0.010 \\
8 & Same & 0.246 & low & 0.998 & 0.000 & 0.002 \\
9 & Same & 0.412 & low & 0.984 & 0.003 & 0.013 \\
10 & Tabular summary: avg price \& totals & 0.303 & low & 0.960 & 0.031 & 0.009 \\
11 & Same table request & 0.563 & low & 0.994 & 0.000 & 0.005 \\
12 & Same & 0.733 & low & 0.958 & 0.001 & 0.042 \\
13 & Same & 0.570 & low & 0.993 & 0.000 & 0.007 \\
14 & Same & 0.674 & low & 0.993 & 0.003 & 0.004 \\
15 & Same & 0.594 & low & 0.994 & 0.003 & 0.003 \\
\bottomrule
\end{tabularx}
\end{table}

\begin{table}[H]
\centering
\caption{BFCL trading episode (\texttt{multi\_turn\_base\_102}): all steps in the evaluated slice (GPT-OSS probes). Gold risk = heuristic projection used for expected tier.}
\label{tab: appendix_bfcl_trading_full}
\begin{tabular}{@{}lccccccc@{}}
\toprule
Step & $p_{\mathrm{tool}}$ & Tier & $p_{\mathrm{low}}$ & $p_{\mathrm{med}}$ & $p_{\mathrm{high}}$ & Gold risk & Expected tool \\
\midrule
0 & 0.997 & high & 0.014 & 0.002 & 0.984 & high & place\_order \\
1 & 0.957 & high & 0.327 & 0.005 & 0.668 & low & get\_order\_details \\
2 & 0.969 & high & 0.175 & 0.013 & 0.812 & medium & cancel\_order \\
3 & 0.767 & high & 0.160 & 0.003 & 0.837 & low & get\_account\_info \\
4 & 0.820 & high & 0.017 & 0.000 & 0.983 & medium & create\_ticket \\
\bottomrule
\end{tabular}
\end{table}
}

\clearpage

\end{document}